# Experiments in Agentic AI for Science

*Judy Fox, Geoffrey Fox.*

*School of Data Science, Department of Computer Science, and Biocomplexity Institute, University of Virginia, Charlottesville, VA, USA. Emails: ckw9mp@virginia.edu and vxj6mb@virginia.edu*

**Abstract**

This paper details two novel frameworks for developing autonomous, agentic AI in scientific workflows. Both systems leverage a hybrid "Local Body, Remote Brain" architecture via Google Colab, utilizing Python-based local orchestrators to invoke large language model (LLM) cloud backends. The first agent, DeepTS/DeepCollector, automates the large-scale curation, extraction, and deduplication of time-series datasets. The second, DeepScribe, is an autonomous presentation analyzer that converts visually dense, mathematically complex physics lectures into structured scientific reports. Through practical systems engineering—such as granular attribute extraction (Cellular RAG), remote data inspection, and distributed concurrency controls—we demonstrate how agentic AI can overcome the context and reasoning limitations of current state-of-the-art systems to rigorously support scientific workflows. Finally, we outline a generalization of DeepTS to support deep knowledge graphs and discuss the application of this conceptual approach to high-energy physics (DeepQCD).

## 1 Introduction

### 1.1 Deep Collector Introduction: The Cataloging Function of DeepTS

The architecture of our proposed DeepTS LLM-based assistant (detailed in Section 2, Figure 1) [1], [2] is designed to manage, automate, and enhance the time series Commons via its DeepCollector cataloging module. Our approach builds upon the Universal Deep Research (UDR) architecture developed at NVIDIA [3]–[5]. It operates as a Python application that invokes computationally intensive operations (LLM inference and web search) on a remote cloud backend. The orchestrator runs flexibly on either a local workstation or Google Colab.

DeepTS utilizes the LlamaIndex [6] library to manage its core Retrieval-Augmented Generation (RAG) pipeline [7]–[9]. It leverages the OpenAI API Specification [10] as the standard wire protocol for LLM interactions. Furthermore, LlamaIndex [11] supports the Model Context Protocol (MCP)[12] for interacting with external tools; while the current implementation does not use it, future iterations of DeepTS will adopt MCP. We currently use Google Gemini as it demonstrated strong reasoning performance during initial prototyping, but the framework is model-agnostic and easily adaptable to other foundation models.

The design was initiated with around a hundred manual prompts, which revealed that LLMs can analyze individual sites, but the integration of the many datasets and model sources required an agentic AI approach. This was able to integrate the results of many inferences and led to our initial repository of 730 datasets. The analysis of most sites is accomplished with a **cellular RAG**. This approach increases the reliability of LLM results by prompting separately for the values of the features for each dataset, and recording a value and confidence for each cell. A few sites need a specialized Harvester which uses very site-specific tools, such as those available for the UC Irvine Machine Learning repository [13]. Future work entails transitioning DeepTS to a production-quality framework, as shown in Fig. 2. This has a straightforward extension to catalog models and benchmarks and to increase dataset coverage, which is currently from the model development community seen in [14]. The choice of important features discovered by the cellular RAG will be evolved, and further, we intend a federated MCP-based Harvester interface allowing DeepTS to be built on top of distributed repositories. We will use Croissant [15] to record the meta-features. A modern user interface will be implemented using React [16] and include a Leaderboard that records model performance and dataset usage.

### 1.2 DeepCollector System Overview and Objective

The rapid proliferation of time-series data across decentralized repositories, academic literature, and pre-print servers creates a significant bottleneck for data curation. DeepCollector is an autonomous, multi-agent Retrieval-Augmented Generation (RAG) system designed to dynamically harvest, synthesize, and structure this fragmented information into a canonical, deduplicated Knowledge Base.

Unlike traditional web scrapers that rely on static DOM parsing, DeepCollector utilizes Large Language Models (LLMs) to perform semantic reasoning over unstructured texts (e.g., arXiv PDFs, GitHub READMEs, web documentation). It extracts highly specific cellular data points—such as variable counts, spatial-temporal frequencies, and source URLs—while probabilistically measuring its own extraction confidence.

### 1.3 The Hybrid "Thin Agent" Architecture

The DeepScribe for Science framework represents a shift from traditional monolithic software to a hybrid agentic pipeline. It addresses the complex problem of converting visually dense physics lectures into formatted academic reports by distributing responsibilities between a local orchestrator and a remote cloud backend. Development of this pipeline utilized an LLM-assisted programming methodology. To accommodate context window limitations, the developer engaged in highly modular iteration, exchanging isolated functions and error logs rather than the full codebase. The pipeline leverages large language models for reasoning and structural formatting, specifically utilizing their capabilities in video processing (ffmpeg) and LaTeX generation. Operational boundaries were navigated through careful session management to accommodate daily rate limits on LLM API invocations. Typical results are demonstrated on the 41 available talks at GenAI25 [17].

#### 1.3.1 The "Hands-Free" Operational Model

The system is designed for "Zero-Touch" autonomy. A user simply provides a Video URL or PDF path and initiates the run. The Colab script (The Agent) then autonomously manages a 45–60 minute lifecycle involving 10 distinct phases, error recovery, and document compilation without requiring any human interaction. Several of the steps were first developed by high-touch human-initiated processes.

#### 1.3.2 Remote Brain, Local Body

- **The Colab Orchestrator (The Agent):** The Python runtime in Colab acts as the "Body." It handles the logistical heavy lifting—downloading 4K video files, chunking audio with ffmpeg, rasterizing PDFs with poppler, and managing file state—without requiring expensive GPUs. It ensures robustness via a **"Model Hopper,"** a failover system that automatically switches models (e.g., from Gemini 3 to 2.5) if API rate limits are hit, preventing the job from crashing.
- **The Gemini Cloud (The Brain):** The cognitive reasoning is offloaded to Google's Vertex AI. The Agent streams visual frames and context to **Gemini 3 Pro** (for reasoning) and **Deep Research Pro** (for web verification), leveraging models with trillions of parameters that could never run locally.

#### 1.3.3 Comparison with State-of-the-Art (SOTA)

Converting visually dense scientific lectures into formatted academic reports highlights several limitations in current state-of-the-art (SOTA) multimodal AI pipelines:

- **Audio-First Transcribers:** Standard transcription pipelines (e.g., Whisper [18]) excel at capturing spoken words but struggle significantly when scientific context relies on visual data (e.g., a speaker silently pointing to a Hamiltonian equation on a blackboard). DeepScribe actively correlates audio transcripts with extracted visual mathematics.
- **Native Long-Context Video LLMs:** While frontier models (e.g., Gemini 1.5 Pro [19] , GPT-4o [20]) can natively ingest hour-long videos, passively passing continuous video directly into the context window often results in a degradation of extraction precision. These models can hallucinate complex derivations if they lack external verification mechanisms.
- **The DeepScribe Innovation:** DeepScribe circumvents monolithic inference limitations through its **Bifurcated Ingestion Engine** and **Agentic Deep Synthesis**. It actively chunks and filters the timeline locally rather than passively watching it. Furthermore, by pausing transcription to integrate Deep Research Pro, it bridges the gap between a simple transcript and a peer-reviewed academic appendix grounded in current literature.

## 2 Case Study 1: DeepCollector for Time-Series Curation

### 2.1 DeepCollector Software Architecture

The architecture is designed as a fault-tolerant, distributed state machine capable of executing highly parallelized network I/O and LLM inference while maintaining strict data integrity. The end-to-end pipeline, illustrating the interplay between the agentic orchestrator and the persistence layer, is depicted in Figure 1.

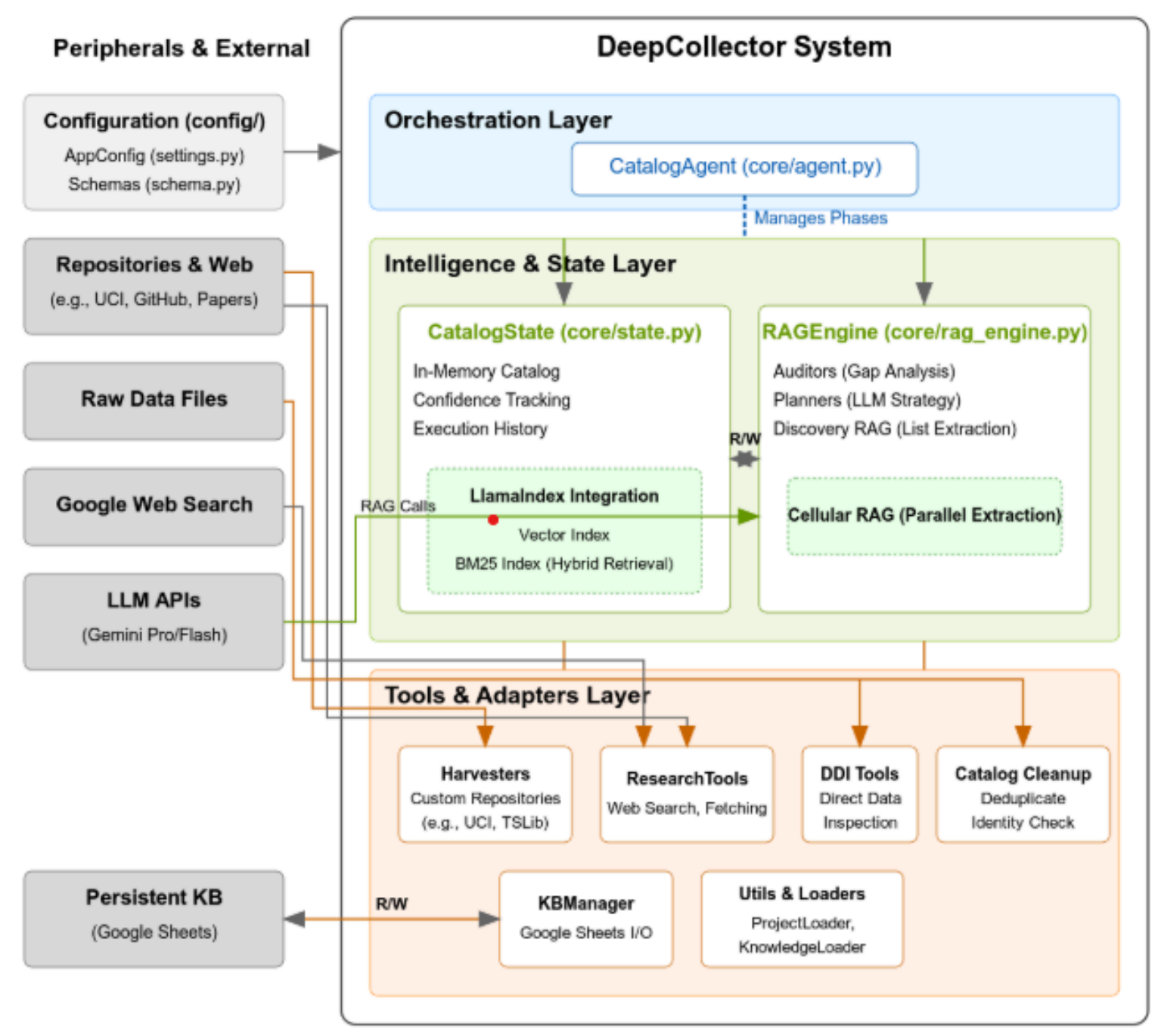

Figure 1: **Software Architecture of the Prototype DeepTS**

#### 2.1.1 The Multi-Agent Orchestration Layer

The system partitions cognitive load across specialized LLM roles:

- **The Planner (Reasoning):** Acts as the strategic node. It analyzes the current state of the catalog, identifies missing foundational datasets, and generates targeted search queries to fetch external grounding context.
- **The Synthesizer (Discovery):** Ingests vectorized documents (via LlamaIndex and BM25) and applies entity taxonomy rules to extract raw candidate records, distinguishing between leaf datasets, parent collections, and synthetic generators.
- **The Extractor (Cellular RAG):** Operates on a highly parallelized thread pool. It targets

low-confidence cells within the Knowledge Base, executing multi-query hybrid retrieval to pinpoint missing variables or URLs. It utilizes self-reflection prompts to resolve discrepancies (e.g., confusing time steps with spatial locations).

### 2.1.2 Hybrid Entity Resolution Engine

To consolidate canonical records from noisy variants (e.g., merging "M4_Hourly" and "M4 Competition Dataset"), the system employs a zero-trust Entity Resolution Engine.

- **Symmetric Difference Discriminators:** Uses deterministic lexical analysis to hard-block the merging of mutually exclusive temporal or spatial variants.
- **LLM Arbitration:** For fuzzy matches, the engine evaluates the complete multi-dimensional state of two candidate records (URL equivalence, dimensionality, descriptions) to issue a definitive boolean merge verdict.
- **Hash Caching:** Verdicts are cached using deterministic tuple keys to prevent redundant compute and mitigate hash collision vulnerabilities.

### 2.1.3 The Distributed Persistence Layer

The persistent Knowledge Base is hosted on a cloud-native relational grid (Google Sheets), abstracted via pandas and gspread.

- **Atomic Title-Based Mutex:** To prevent TOCTOU (Time-of-Check to Time-of-Use) race conditions in parallel executions, the system achieves true mutual exclusion by injecting the epoch timestamp and job ID directly into a transient worksheet's title. This leverages the backend's strict global uniqueness constraint to block simultaneous write access.
- **Vectorized Sanitization:** A robust data ingestion pipeline flattens list/dictionary artifacts natively emitted by LLMs, sanitizing formula injections and bounding-box overflows before applying asynchronous write-backs.

## 2.2 Related Agentic Work and System Positioning

To contextualize DeepCollector's contributions, it is necessary to contrast its design with existing paradigms in autonomous AI agents, Retrieval-Augmented Generation (RAG) [21], and unstructured data extraction.

- **Generalist Autonomous Agents:** Traditional agentic frameworks—such as AutoGPT [22], Devin [23], and LangGraph [24]—excel at open-ended, sequential task solving (e.g., autonomous software engineering, planning itineraries) [25]. However, when applied to large-scale data curation, these unbounded systems frequently suffer from "state drift," infinite hallucination loops [26], and a failure to adhere to strict relational database schemas over prolonged execution windows. **DeepCollector Innovation:** We overcome this by constraining LLM randomness within a rigid, deterministic state-machine architecture (Bootstrapping → Discovery → Grounding → Extraction → Repair), drawing on constrained decoding principles [27].
- **Academic and Literature Review Agents:** Existing research agents, such as STORM [28], Elicit [29], and SciSpace [30], successfully utilize RAG for high-level literature synthesis. Yet, they rarely verify the physical existence of underlying data files. **DeepCollector Innovation:** Our framework advances this paradigm by employing a "Pre-Flight Crawler" that actively hunts down supplemental code repositories (e.g., GitHub, Zenodo) to verify textual claims via physical file inspection.
- **Semantic Web Scrapers:** Modern semantic scrapers, such as ScrapeGraphAI [31], and those evaluated in the Mind2Web benchmark [32], rely on LLMs to translate HTML/DOM structures into JSON outputs. **DeepCollector Innovation:** We transcend text-based DOM parsing by implementing Direct Data Inspection (DDI), enabling the agent to bypass metadata descriptions entirely and physically inspect raw data payloads over the network.

### 2.3 Unique Architectural Innovations

DeepCollector introduces several novel engineering paradigms for robust agentic workflows, particularly when operating in stateless, serverless (e.g., Google Colab), or highly parallelized environments.

#### 2.3.1 Distributed Locking via Cloud Spreadsheets (Title-Based Mutex)

Coordinating parallel multi-agent executions typically requires dedicated infrastructure (e.g., Redis, Apache ZooKeeper) to manage distributed locking and prevent Time-of-Check to Time-of-Use (TOCTOU) race conditions [33]. To minimize infrastructure overhead, DeepCollector utilizes the strict global uniqueness constraints of Google Sheets tab names as a lightweight mechanism for mutual exclusion. By injecting the epoch timestamp and Job ID into a transient worksheet title, the system natively leverages the Google backend as a robust atomic lock, complete with a garbage collection mechanism for orphaned locks to ensure fault tolerance.

#### 2.3.2 "Cellular RAG" with Granular Confidence Tracking

Rather than prompting an LLM to generate an entire JSON record or paragraph—which risks compounding hallucinations [26]—DeepCollector executes "Cellular RAG." It treats every individual cell in the target schema as its own isolated extraction task (e.g., querying specifically for Num Time Points). Every cell maintains its own tuple of (Value, Confidence, Rationale, Source URL). Relying on LLM uncertainty estimation [34], if the extraction confidence falls below a defined CONFIDENCE_LOCK_THRESHOLD (e.g., 0.95), the system does not guess; it flags the cell as [missing] and queues it for a highly targeted "Repair Sweep" in subsequent iterations.

#### 2.3.3 Direct Data Inspection (DDI) via HTTP Range Peeking

LLMs cannot natively count rows in a 50GB .csv or .zip file, nor can a serverless agent afford to download massive payloads into memory. Furthermore, academic abstracts frequently misreport their own data dimensions. The DDI module circumvents this by utilizing HTTP Range headers (bytes=0-1048575) to stream-download only the first megabyte of a target data file [35]. It then uses deterministic libraries (pandas) to parse the snippet in memory, perfectly verifying variables and dimensional shapes to mathematically ground the claims extracted by the RAG engine.

#### 2.3.4 The Hybrid Zero-Trust Entity Resolution Engine

Entity resolution (ER) and master data management across thousands of noisy, web-scraped records is prohibitively expensive and error-prone for LLMs acting alone [36]. DeepCollector uses a hybrid approach: fast, deterministic "Symmetric Difference Discriminators" instantly block impossible merges (e.g., recognizing that "Hourly" and "Daily", or geographic locations like "Germany" and "France", are mutually exclusive). Only when records are "fuzzy matches" does it invoke an LLM Arbitrator to evaluate the full multi-dimensional state. The Arbitrator caches these tuple-based verdicts to drastically reduce API costs.

#### 2.3.5 Discrepancy Arbitration and Self-Healing Cascades

When web sources conflict (e.g., an academic abstract claims 50 variables, but the GitHub repository documentation shows 45), standard LLMs often hallucinate an average [26]. DeepCollector utilizes explicit Arbitration Prompts, forcing the LLM to prioritize structural repository files over high-level abstracts and explicitly log its decision-making rationale. Furthermore, if an LLM endpoint hangs or hits a rate limit (HTTP 429), a thread-safe "guillotine" timeout catches it, applies a penalty strike, and seamlessly cascades the query to a fallback model—employing principles of verbal reinforcement and self-correction [37].

### 2.4 Extensibility and Generalizability of the Methodology

While the current instantiation of DeepCollector is optimized for cataloging time-series datasets, its underlying methodology—Autonomous Cellular Curation—is fundamentally domain-agnostic. The "Freeze and Focus" modularity (separating the stable orchestration engine from the mutable JSON schemas) allows the framework to be generalized to virtually any domain requiring the curation of fragmented, unstructured data into a canonical structure [38].

#### 2.4.1 Expanding to the Scientific Knowledge Graphs (The Relational Triad)

Tracking datasets is only one piece of the AI development lifecycle. DeepCollector's architecture is highly extensible and aligns perfectly with the goals of working groups standardizing AI and HPC workflows. The system can seamlessly generalize to map the entire ecosystem by shifting from a flat dataset catalog to an active knowledge graph, structured as a Relational Triad (Datasets ←> Models ←> Benchmarks):

- **The Models Table (Architecture Tracking):** The Harvester and Cellular RAG agents can be repurposed to ingest academic papers and code repositories. By adapting the DDI module to inspect Hugging Face repository structures [39] (e.g., physically reading config.json or .safetensors headers), the system can deterministically extract model traits.
  - *Fields:* Model Architecture (e.g., Transformer, Mamba, Diffusion), Parameter Count, Context Window, Primary Framework (PyTorch/JAX), Hardware Requirements, and Checkpoint URLs.
- **The Benchmarks/Evaluation Table (Automated Leaderboards):** The core challenge in AI research is tracking which model performed best on which dataset under specific conditions. The Multi-Query Cellular RAG agent can be directed at PDF evaluation tables in arXiv pre-prints to extract evaluation metrics. This builds an intersection table mapping DatasetID to ModelID, generating a living, automated leaderboard of AI research.
  - *Fields:* Evaluation Metric (e.g., MSE, MAE, CRPS), Zero-Shot vs. Fine-Tuned, Hyperparameter configuration, Hardware utilized (e.g., DGX A100), and Training Compute (FLOPs).
- **Required Architectural Adjustments:** To facilitate this extension, minor architectural tuning is needed:
  - *DDI Expansion:* The current inspection tools would need to expand beyond parsing .parquet and .csv data files to inspecting .json files from model cards to automatically extract parameter counts and tensor shapes.
  - *Table Extraction Logic:* The Pre-Flight Crawler requires enhanced multi-modal capabilities or dedicated OCR logic to parse performance metrics out of complex PDF tables in academic papers, converting them into structured JSON for the LLM Arbitrator to evaluate.
  - *Entity Resolution for Models:* The Entity Resolution Engine needs new discriminators. For example, it must recognize instruction-tuning as a hard boundary, ensuring that *Llama-3-8B-Instruct* is not erroneously merged with *Llama-3-8B-Base*.

### 2.4.2 High-Performance Computing (HPC) Workloads and Code Translation

While initially validated on time-series datasets, DeepCollector's Autonomous Cellular Curation framework is fundamentally domain-agnostic and uniquely positioned to manage the strict constraints of High-Performance Computing (HPC). Adapting legacy scientific codes to modern, energy-constrained heterogeneous accelerators requires complex dependency resolution. DeepCollector's Direct Data Inspection (DDI) module can be extended beyond dataset files to physically inspect distributed version control systems (e.g., GitHub) and parse C++, SYCL, and CUDA files. To handle the proliferation of variant legacy codes, the Entity Resolution Engine can replace purely semantic LLM arbitration with deterministic Abstract Syntax Tree (AST) hashing [40], mathematically guaranteeing the identification of algorithmic code clones.

Furthermore, the framework's Harvesters can be adapted to programmatically extract structural meta-features from LLVM Intermediate Representation (IR) passes [41]. Extracting hardware-specific metadata—such as compute-to-memory ratios—provides deterministic inputs for pre-execution energy footprint modeling [42]. Ultimately, applying Cellular RAG to HPC metadata lays the groundwork for standardizing scientific benchmark ontologies [43] and curating high-energy physics hardware-synthesis workflows [44]. This automated metadata extraction serves as the foundational data orchestration layer for the Department of Energy's Exascale ecosystem. Ultimately, combining this capacity for automated code inspection with the dynamic knowledge graph structures outlined above provides the foundation for autonomous, domain-specific AI agents—a paradigm we formalize as DeepQCD.

## 3 Case Study 2: DeepScribe for Multimodal Science

The DeepScribe system is an autonomous presentation analyzer engineered to operate under a fully autonomous ("zero-touch") paradigm. The current iteration is open-source and available via Google Colab [45].

### 3.1 Architecture Breakdown: The 10 Components

The pipeline functions as a linear state machine with 10 distinct components (cells) (Initialization + Phases 0–9).

**Initialization: The Model Hopper**

The system initializes the API client with a self-healing logic. If the primary experimental model (gemini-3-pro-preview) fails due to high traffic (503) or rate limits (429), the Agent automatically reroutes requests to the stable backup (gemini-2.5-pro) and continues execution. Gemini-2.5-pro is used in the Phase 7, and gemini-3-pro-preview is used in the more critical initial Scribe phase. Note that at the moment, there is a daily cap on the number of gemini-3-pro-preview invocations. During execution, DeepScribe typically invokes the primary reasoning model (e.g., Gemini 3 Pro) and the compilation model (e.g., Gemini 2.5 Pro) once per defined "event" (such as a slide transition), with cloud execution times averaging 10-20 seconds per call. The Deep Research agent is invoked once per event to gather external context, executing for approximately 10 minutes.

### 3.1.1 Phase 0 & 1: The Bifurcated Ingestion Engine

The Agent automatically detects the input type and routes the data through specialized logic:

- **Case A (Video/Blackboard):** The Agent sends video chunks to **Gemini 2.0 Flash** (Cloud) to build a timeline. It employs **Smart Gap Logic**: if it detects silence >120 seconds (indicative of blackboard writing), it inserts a "Blackboard Event" to force visual analysis of that segment.
- **Case B (PureSlide/PDF):** The Agent uses **Header-Locked Fusion** (Local). This algorithm

compares perceptual hashes of slide headers to collapse "build-up" animations (where bullet points appear one by one) into a single image, ensuring the AI only analyzes the complete slide.

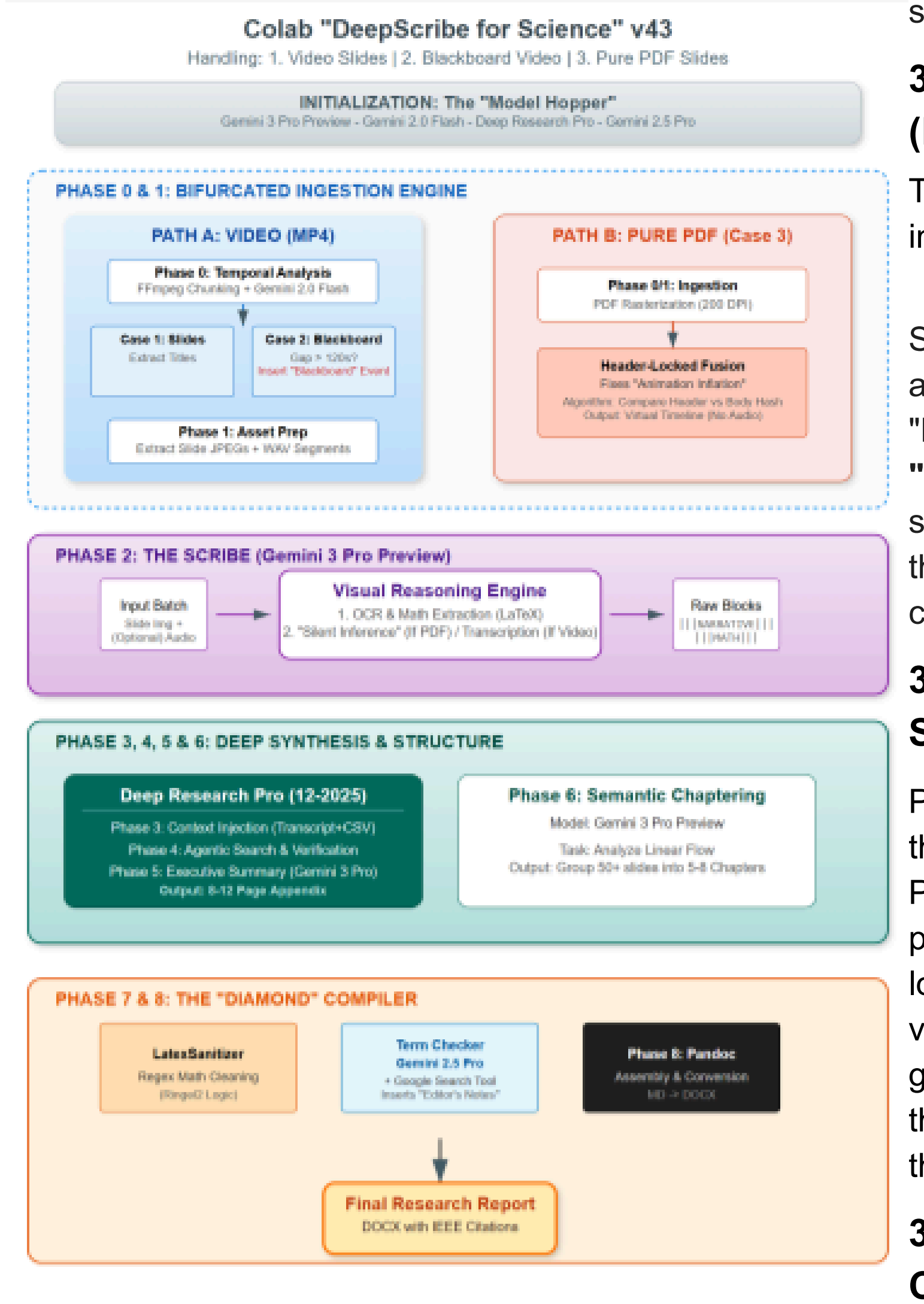

Figure 2: DeepScribe architecture

In

### 3.1.2 Phase 2: The Scribe (Multimodal Cloud Inference)

The Agent uploads batches of images and audio to the Cloud.

- **Gemini 3 Pro** acts as the Scribe. In "Video Mode," it correlates audio with visual equations. In "PureSlide Mode," it performs **"Silent Inference,"** inferring the speaker's narrative based solely on the visual density of the slide content.

### 3.1.3 Phase 3, 4 & 5: Deep Synthesis (Agentic Research)

Phase 3 (Context Injection) compiles the raw transcript and slide structure. Phase 4 (Deep Research), the agent performs recursive web searches to locate relevant source literature, verify physics definitions, and generate a comprehensive appendix that situates the presentation within the current scientific literature.

### 3.1.4 Phase 6: Semantic Chaptering

Gemini 3 Pro reads the linear transcript and groups the 50+ slides into 4–8 logical chapters (e.g., "Introduction," "Lattice Formulation," "Results"), providing a narrative structure to the report.

### 3.1.5 Phase 7 & 8: The "Diamond" Compiler (Local Assembly)

The final assembly is handled locally to ensure formatting safety.

- **LatexSanitizer:** A Regex engine cleans the AI's LaTeX output.
- **Term Checker:** Gemini 2.5 Pro (with Google Search) is called to define complex jargon ("Editor's Notes").
- **Final Output:** The Agent invokes Pandoc to compile the Markdown, Images, and LaTeX into a polished DOCX file, saved directly to Google Drive.

## 4 System Evaluation & Telemetry: Compute and API Utilization

The DeepCollector pipeline utilizes a multi-model, agentic architecture to balance advanced reasoning capabilities with high-throughput data extraction. Execution relies on a dual-cascade model configuration: a reasoning-heavy model (gemini-3.1-pro-preview) is deployed for workflow planning, heuristic discovery, and complex search query generation, while a highly concurrent, lower-latency model (gemini-3-flash-preview) handles the Cellular RAG metadata extraction and structuring.

To evaluate the scaling characteristics of the pipeline, resource utilization was profiled across a spectrum of dataset collections, ranging from single-domain benchmarks to massive, multi-domain archives. *Note: While this evaluation focuses on computational overhead, API latency, and scaling dynamics, a rigorous quantitative assessment of extraction accuracy (e.g., precision and recall) and hallucination rates against a human-annotated baseline remains a priority for future work.* Because the system leverages asynchronous execution during its extraction phase (parallel Cellular RAG batches), the cumulative CPU computation time often heavily exceeds the actual wall-clock execution time.

**Table 1: Workflow Execution & CPU Utilization by Job Size**

| Project (Job Size) | Datasets Cataloged | Wall-Clock Time | Cumulative CPU Time* | Time of Deep Research |
|---|---|---|---|---|
| **M5 (Small)** | 5 | 23.7 mins (1,422s) | 0.39 hours (1,440s) | 600s |
| **M6 (Small)** | 8 | 54.3 mins (3,259s) | 1.26 hours (4,554s) | 600s |
| **TEMPO (Medium)** | 19 | 36.0 mins (2,158s) | 0.63 hours (2,277s) | 600s |
| **FLOWSTATE (Medium)** | 20 | 103.3 mins (6,202s) | 3.67 hours (13,202s) | 600s |
| **KRONOS (Large)** | 20 | 110.0 mins (6,601s) | 2.92 hours (10,557s) | 600s |
| **LOTSA (Massive)** | 58 | 314.0 mins (18,844s) | 12.77 hours (45,970s) | 600s |

**Cumulative CPU time represents the total time the LLM spent generating tokens across all concurrent threads.*

**API Call Volume and Latency Dynamics**

The latency and volume of API calls vary significantly depending on the role of the model within the workflow. They also vary due to experimental status of some of models. Note our code runs on Colab but search and LLM are invoked externally on Google Cloud so timing includes wait times in transit and due to Cloud loading time.

**Table 2: LLM Call Volume & Average Latency by Model Type**

| Project | Extraction (Flash) | Planning (Pro) | Search/Grounding (Pro) |
|---|---|---|---|
| **M5** | 72 calls (13.5s avg) | 4 calls (12.8s avg) | 11 calls (38.1s avg) |
| **M6** | 184 calls (15.8s avg) | 4 calls (14.0s avg) | 31 calls (51.4s avg) |
| **TEMPO** | 162 calls (6.9s avg) | 4 calls (18.5s avg) | 23 calls (47.4s avg) |
| **FLOWSTATE** | 356 calls (30.4s avg) | 4 calls (18.5s avg) | 45 calls (53.2s avg) |
| **KRONOS** | 353 calls (18.7s avg) | 4 calls (20.1s avg) | 65 calls (59.7s avg) |
| **LOTSA** | 1,141 calls (33.8s avg) | 4 calls (37.5s avg) | 164 calls (44.5s avg) |
| **Aggregate Average** | 2,268 calls (26.9s avg) | 24 calls (20.2s avg) | 339 calls (49.2s avg) |

- **The Deep Research Baseline:** Across all job sizes, the Deep Research agent—utilized during Phase 0 Bootstrapping—consistently hit a quota-limited ceiling of exactly 600 seconds (10 minutes) per run. Rather than crashing, the agent is designed to gracefully halt exploration and synthesize its findings at this hard limit, successfully yielding highly relevant foundational datasets (e.g., 2 datasets for M5, 17 for LOTSA) to seed the subsequent discovery loops. As an experimental prototype API, the underlying Deep Research tool exhibits variable behavior; this explicit timeout enforcement ensures pipeline stability and prevents indefinite execution holds.

- **Extraction Overhead (Flash):** The Phase 3 Extraction is the most compute-intensive segment of the pipeline, handling the bulk of the transaction volume. Across the six profiled runs, the gemini-3-flash-preview model accounted for 2,268 individual calls. Average latency scales with job size (from 6.9s on TEMPO up to 33.8s on LOTSA), likely due to the increased context window payloads generated by the massive accumulation of Web and RAG documents in larger jobs, as well as momentary rate-limiting backoffs.
- **Search Dependency (Pro + Search Tool):** Grounding the datasets via web retrieval yields the highest per-call latency (averaging 49.2 seconds globally). This elevated latency is expected, as it incorporates the physical network overhead of executing live Google searches, retrieving external HTML, and summarizing the results before returning the payload to the agent.
- **Strategic Planning (Pro):** The pure reasoning tasks handled by the gemini-3.1-pro-preview model required exactly 4 calls per job to orchestrate the search strategies. Because this is a closed-book task evaluating a highly predictable context string, its latency remained relatively stable and efficient (averaging 20.2s).

## 5 Generalizing the Architecture: DeepKG & DeepQCD

While the initial instantiations of our Agents architecture—DeepTS (including DeepCollector) for computer science workflows and DeepScribe for multimodal scientific presentation analysis—have proven highly effective in particular applications, their underlying multi-agent methodologies are fundamentally domain-agnostic. To transition from domain-specific tools to generalized frameworks for scientific data orchestration, we abstract our architecture into DeepKG **(Deep Knowledge Graph)**.

### 5.1 DeepKG: Domain Agnosticism via Active Knowledge Graphs

DeepKG represents a generalized agentic framework where nodes and links are defined by arbitrary domain ontologies. By decoupling the orchestration mechanics from the specific subject matter, the TPCAgents engine can operate interchangeably. Whether the active ontology represents "Time Series Machine Learning," "Quantum Mechanics," or "Climate Modeling," the DeepKG engine functions identically to build complex relational graphs from fragmented data. The architecture enforces strict scientific rigor through four core pillars:

1. **Domain Agnosticism via Ontologies**: We define DeepKG as a flexible set of ontologies (the formal rules of any domain) and the mathematical/logical links between cells, rendering the system universally generalizable.
2. **Provenance and Trust (Anti-Hallucination):** To combat the "black box" nature of LLMs, DeepKG explicitly ensures that every single node, cell, and link is associated with a grounding context. If the AI proposes an equation, researchers can trace it back to the exact dataset or paragraph in the original source paper. This addresses strict standards for reproducibility and mirrors the provenance-tracking priorities of the InferA framework [46]. This was developed by Los Alamos National Laboratory (LANL) researchers with an Argonne National Laboratory collaborator. InferA's source code is released at [47]. Like DeepKG, InferA utilizes a supervisor agent coordinating specialized sub-agents to handle scalable analysis of terabyte-scale cosmological ensemble data without fully ingesting it into memory, rigorously tracking provenance to guarantee scientific reproducibility.This methodology also provides an interesting comparison to other work by LANL detailing an autonomous agentic framework for scientific discovery [48].

3. **Uncertainty Quantification (UQ):** The system natively assigns a probabilistic confidence level to every node and edge. By explicitly quantifying its uncertainty, the AI agent is aware of exactly when it is estimating or guessing.
4. **Active Agentic Behavior:** Standard knowledge graphs are passive. DeepKG utilizes "robust iterative knowledge searches," actively curating its own graph. When it detects low-confidence nodes, it autonomously launches targeted web or database searches specifically designed to boost certainty, establish missing links, and discover new nodes.

### 5.2 Grounding DeepKG in Physics: The DeepQCD Paradigm

To demonstrate the generalizability of DeepKG, we map the architecture directly into the domain of high-energy nuclear physics to create DeepQCD: an autonomous agentic framework designed to resolve the petabyte-scale data synthesis bottlenecks of modern nuclear and particle physics analysis. The transition from DeepTS to DeepQCD unfolds across four agentic phases:

### 5.2.1 Constructing the Nodes (The Knowledge Graph)

Both systems use an AI agent to ingest massive amounts of fragmented, multimodal information and link concepts to reveal new relationships. In the computer science domain, DeepTS reads AI papers, code repositories, and databases to construct a graph linking [Datasets] → [ML Architectures] and [Hardware Configurations] → [Performance Benchmarks].

By swapping the ontology, DeepQCD acts as a domain-expert agent. It ingests unstructured arXiv preprints, massive Lattice QCD databases, and Jefferson Lab experimental data. It constructs a physics-based DeepKG linking [Theoretical Concepts] → [LQCD Predictions] and [Mathematical and Physical Symmetries] → [Phenomenology] → [Experimental Observables]. This methodology parallels ArgoLOOM (Argonne-based system of Linked Oracles for Observables and Models), a demonstrator platform that utilizes Python steering scripts interfacing with LLMs and a curated theory knowledge base to orchestrate multi-domain physics modeling from quarks to cosmos [49].

### 5.2.2 Identifying the Gap (The "Trigger" for Discovery)

Because the agent maintains a continuous, multidimensional map of reality, it can autonomously detect physical tension or anomalies. Where DeepTS might notice a performance gap (e.g., "Model X is highly accurate theoretically, but no one has optimized it for Hardware Y on Dataset Z"), DeepQCD scans its physics graph to identify theoretical contradictions. For example: "Theoretical paper X predicts a specific string topology at low x, but the extracted Compton Form Factors from the experimental data do not match this prediction." Recent evaluations, such as the SciCrafter benchmark, identify "Knowledge Gap Identification" as the defining bottleneck for autonomous science; DeepQCD targets this by prompting its reasoning orchestrator to actively identify empirical-theoretical anomalies [50].

### 5.2.3 Closing the Loop (Autonomous Action & Hypothesis Generation)

Guided by the Perception-Language-Action-Discovery (PLAD) closed-loop paradigm [51], the proposed DeepQCD agent is designed to take autonomous action rather than merely pointing out gaps to a human physicist. It aims to hypothesize physical mechanisms to explain anomalies. Because standard LLMs are prone to mathematical hallucinations in quantum field theories, DeepQCD intends to translate these constraints to an external symbolic framework: an Agent-Guided Multi-Objective Neuro-Symbolic Regression (MO-NSR) module [52], [53]. By guiding the symbolic generator with neural priors, the framework's goal is to generate closed-form, human-readable mathematical equations that strictly obey Lorentz invariance and exact forward limits.

The proposed conceptual pipeline would then automatically author Python/C++ scripts to inject these novel equations into a physics event generator (DEEPGen), producing simulated "pseudo-data" to test whether its hypothesis mathematically resolves the tension. This theoretical capacity for automated experimental execution is supported by recent breakthroughs like the Just Furnish Context (JFC) framework, which demonstrated that AI agents can autonomously execute complex experimental high-energy physics analyses [54].

### 5.2.4 The Final Output (The Deliverable)

Finally, DeepQCD evaluates the success of its experiment. Just as DeepTS outputs recommended algorithmic pipelines and deployment strategies, DeepQCD recommends entirely new physical theories. It acts as an autonomous research orchestrator, outputting specific, actionable directives for experimental validation, such as recommending the exact kinematic configurations required for upcoming Electron-Ion Collider (EIC) measurements.

Leveraging the advanced LaTeX compilation capabilities inherited from the DeepScribe module, DeepQCD perfectly formats the complex quantum equations, generates the cross-section plots, outlines the exact kinematic bins needed for the EIC, and outputs a complete scientific paper ready for physicist peer review. Ultimately, frameworks like DeepKG and DeepQCD ensure rigorous adherence to standardized evaluation methodologies, such as the EAIRA principles developed at Argonne [55], providing a verifiable, data-grounded foundation for autonomous scientific workflows.

## 6 Discussion and Future work

### 6.1 Future Work: GraphRAG and Automated Benchmarking

As we build and use the system, we will evolve the schema for datasets, models, and benchmarks to enhance **recommendation** ability. This will be the major system upgrade shown in Fig. 3, that will need to link the different types of collected data and use this to suggest which models are good choices for new datasets and which are good datasets for pretraining new Foundation models. This requires extending DeepTS to build Knowledge graphs [56], [57]. These will include nodes that represent datasets, models, papers, and concepts (meta-features) extracted by the cellular RAG approach. However, the intelligence of the system will lie in the graph edges that link the nodes with, for example, datasets and models, both linked to concepts through their features, and with benchmarks linking models to datasets. We need to extend the cellular (row-based) RAG to a Graph-based RAG [58]. GraphRAG [59]–[66] was introduced by Microsoft Research in 2024 [67] and naturally extends our Knowledge Base into a Knowledge Graph. This methodology is supported by recent work [68], which demonstrates that LLMs can effectively act as zero-shot model selectors when prompted with these specific data characteristics. TimeGNN [69], using features to derive a graph-based neural network model for time series, can be compared to DeepTS, which maps applications to the best of all available models. We will measure DeepTS's success on new data or models (the cold-start issue) as one metric of success [70].

There are three further enhancements, with the most important being its support of **automated benchmarking** [71]–[73]**.** For this, DeepTS will add the capability of a robust Workflow Orchestrator, managing containerized execution environments, automating model onboarding via LLM agents, and enforcing thoughtful hyperparameter optimization and ablation studies. To address the issue of **choosing a good set of meta-features**, DeepTS will evolve from a passive catalog into an active analysis and orchestration platform [74], [75]. For meta-feature validation, DeepTS will implement a dedicated Meta-Learning Module that leverages GraphRAG to empirically assess feature importance and agentically identify gaps to drive schema evolution. To improve data quality, DeepTS will enforce rigorous traceability by requiring "source grounding," linking every RAG extraction to its origin in the Knowledge Graph. Confidence scores will automatically triage low-certainty data into a Human-in-the-Loop (HITL) workflow for expert (crowd-sourced) review, with automated cross-verification across documentation and raw data [55], [76], [77].

The plan described above is ambitious, and we will adjust our plans if necessary in the GraphRAG and LLM-controlled execution areas. To mitigate the inherent complexity of automated benchmarking, we will adopt a phased approach. Initially, we will prioritize Standardized Semi-Automation by integrating established libraries (e.g., NeuralForecast) and requiring community contributions to adhere to the MLCube standard [78]. DeepTS will validate and orchestrate these submissions using standardized hyperparameter optimization approaches. This ensures immediate reproducibility and a robust Leaderboard while the advanced agentic onboarding capabilities mature. The primary recommendation engine will utilize GraphRAG for deep, contextual reasoning. However, recognizing that GraphRAG is a rapidly evolving technology, we will concurrently implement a robust Feature-Based Similarity Search as a parallel pathway. This approach leverages the structured meta-features extracted by Cellular RAG. We will employ established meta-learning techniques to identify datasets with similar characteristics [79], [80], and recommend models that have empirically demonstrated superior performance on those datasets [72], [74]. This ensures the Commons provides high-quality, data-driven recommendations as the GraphRAG implementation matures.

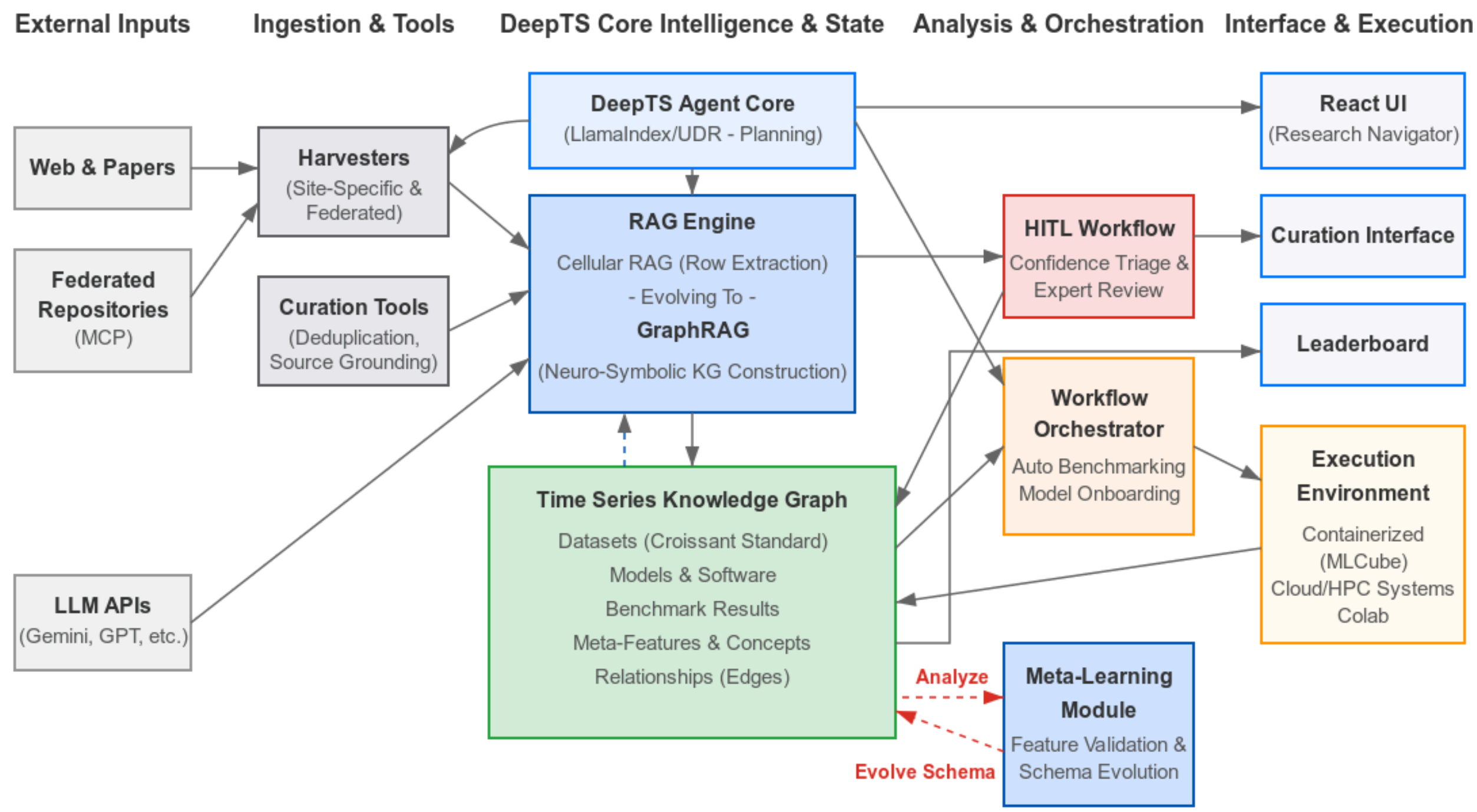

*Fig. 3: DeepTS Target Architecture: An Intelligent, Automated Time Series Commons Navigator*

## 6.2 Deployment Infrastructure and Security

We must deploy proper computational resources to host the many aspects of this project. GitHub will host the initial website [81], open-source codebase, model definitions (Python/Notebooks), and static documentation. Our data strategy emphasizes federation to minimize long-term central storage costs. We will prioritize linking to external repositories via the Federated MCP data interface. For key certified datasets requiring centralized hosting, we will leverage the Open Storage Network [82], aligning with our collaboration with SDSC. The UVA team intends to use this project resource or UVA high performance computing clusters to host the dynamic web portal (production React UI), the DeepTS API, the Knowledge Graph, and the Workflow Orchestrator. We can also use this NSF/UVA resource or Google Colab for executing the models on datasets.

A central computational need is for resources to run the LLM and search tools component of the DeepTS cataloging and recommendation tasks. We will employ a tiered inference strategy. High-throughput tasks (Cellular RAG) will utilize cost-effective models, which today are Gemini Flash. Frontier models (today Gemini Pro) will be reserved for complex reasoning (GraphRAG). AI is improving rapidly, and recent analysis [83]–[86] suggests that equally capable open-source alternatives (perhaps Gemma-based) are available with a time delay of around 6-12 months. As we support open interfaces, we can host the DeepTS backend, where feasible, on open-source LLMs running on NSF-funded cyberinfrastructure and UVA institutional resources, reducing long-term reliance on commercial systems. We are also exploring a purely local implementation using SearXNG [87] and Gemma [88] running on a dedicated 4 A100 80GB workstation.

It is always essential to address cybersecurity issues properly and adopt state-of-the-art approaches to avoid security failures. This is typically defined by the resource provider, but we do need to proactively address the special concerns coming from the use of LLMs [89]–[92]. One screens LLM inputs using tools like Llama Guard [93] and then looks at JSON outputs [94], [95]. The DeepTS Agent Core performs rigorous Schema Validation using Pydantic [96] or a similar tool.

Further, although our Federated Execution Model shifts the primary operational burden of securing community benchmarking runs to the resource providers, the Commons retains critical responsibility for the integrity of the ecosystem, which we address by adopting a zero-trust approach to all externally generated data entering the system. All external inputs—including artifacts submitted for distribution (model code, weights, MLCubes [78], datasets) and execution data submitted to the Leaderboard (results, logs)—will be treated as untrusted, Rigorous security screening of external input will include malware scanning and static analysis (for code/containers), and strict input validation, sanitization, and secure parsing (for results/logs). We are also responsible for the security of our central infrastructure, which we address through robust implementation of hardened sandboxes (such as gvisor [97] or Kata [98]) or the recommended policy of our provider.

### 6.3 Community Adoption

The Time Series Commons requires a high-quality AI assistant, but will only succeed if it creates active community participation. We need to incentivize researchers to both use the Commons and to contribute their data and models, perform benchmarking studies, help in the HITL expert review of LLM results, and adhere to the proposed metadata standards. The three AI consortia, MLCommons, the AI Alliance, and the Trillion Parameter Consortium give us links to the technology community. Initially, we will not just build out the technology, but we will use the DeepTS prototype to create a new time series Foundation model, which improves on Sundial [99], plus new time series model advances in our two application domains. We encourage involvement in the commons by the opportunity of co-authored papers (our latest MLCommons paper [100] has 30 authors from 26 organizations) and awarding of badges (e.g. expert curator) for the HITL activities.

## 7 Conclusions

This paper demonstrates that autonomous, agentic AI frameworks can effectively resolve critical bottlenecks in scientific data curation and multimodal knowledge synthesis. By employing a hybrid "Local Body, Remote Brain" architecture, our systems decouple heavy local orchestrations from cloud-based cognitive reasoning, overcoming the context and hallucination limitations of traditional, monolithic AI models.

Our two primary case studies validate this approach across distinct scientific modalities:

- **DeepCollector** establishes a highly parallelized, self-correcting pipeline capable of autonomously harvesting, deduplicating, and standardizing unstructured time-series data into a mathematically grounded knowledge base.
- **DeepScribe** proves that agentic AI can successfully navigate visually dense, mathematically complex multimodal presentations, correlating audio with visual equations to generate Ph.D.-level, peer-reviewed scientific reports without human intervention.

Crucially, the underlying methodology of "Autonomous Cellular Curation" is domain-agnostic. As formalized through the abstraction of DeepKG (Deep Knowledge Graphs) and its application to high-energy physics via DeepQCD, this architecture provides a blueprint for universal engines of scientific discovery. By actively hunting for empirical-theoretical anomalies and enforcing strict provenance tracking, frameworks like DeepKG lay the groundwork for verifiable, autonomous research orchestration.

Ultimately, these experiments illustrate that agentic AI is no longer just a tool for localized task automation; it is a foundational cyberinfrastructure capable of accelerating reproducible, community-driven scientific inquiry across the Exascale ecosystem.

## Acknowledgements

The work on DeepScribe was motivated by the visit of Geoffrey Fox to the KITP GenAI25 meeting. This research was supported in part by grant no. NSF PHY-2309135 to the Kavli Institute for Theoretical Physics (KITP). Many fruitful interactions with the MLCommons Science Working Group and the AI Alliance are gratefully acknowledged. Discussions with Franck Cappello on AI Science Assistant were important.

## References

[1] MLCommons, University of Virginia, Geoffrey Fox, “Catalog of about 700 Deduplicated Time Series datasets produced by Prototype AI Time Series assistant DeepTS,” 21-Mar-2026. [Online]. Available: https://docs.google.com/spreadsheets/d/1-PuWrHO30E4WPM-rOed03n42gfo5AlEtscKqqtjznA0/edit?usp=sharing. [Accessed: 22-Mar-2026]

[2] Geoffrey Fox, “Benchmarking for HPC AI in a rapidly changing world at Smoky Mountains Computational Sciences and Engineering Conference, August 25 – 28, 2025, Chattanooga, TN,” 26-Aug-2025. [Online]. Available: https://docs.google.com/presentation/d/1aCGjbAmdHG6zw5yFrUmEhqu8gYQcqbDQVsNrkRdCzHM/edit?usp=sharing. [Accessed: 22-Nov-2025]

[3] P. Belcak and P. Molchanov, “Universal Deep Research: Bring your own model and strategy,” *arXiv [cs.AI]*, 29-Aug-2025 [Online]. Available: http://arxiv.org/abs/2509.00244

[4] V. Hsiao, M. Roberts, and L. Smith, “Procedural Knowledge Improves Agentic LLM Workflows,” *arXiv [cs.AI]*, 10-Nov-2025 [Online]. Available: http://arxiv.org/abs/2511.07568

[5] Joshua Berkowitz, “Universal Deep Research: A User-Programmable Deep Research Agent: NVIDIA’s UDR Decouples Strategy From Models And Lets Users Bring Their Own Workflow,” 07-Sep-2025. [Online]. Available: https://joshuaberkowitz.us/blog/papers-7/universal-deep-research-a-user-programmable-deep-research-agent-1040. [Accessed: 24-Nov-2025]

[6] LlamaIndex, “LlamaIndex redefining document workflows with AI agents.” [Online]. Available: https://www.llamaindex.ai/. [Accessed: 22-Nov-2025]

[7] Salesforce, “What Is Retrieval-Augmented Generation (RAG)?” [Online]. Available: https://www.salesforce.com/agentforce/what-is-rag/. [Accessed: 24-Nov-2025]

[8] r/RAGCommunity, “Hybrid Vector-Graph Relational Vector Database For Better Context Engineering

with RAG and Agentic AI," Sep-2024. [Online]. Available: https://www.reddit.com/r/RAGCommunity/comments/1nfqyx3/hybrid_vectorgraph_relational_vector_database_for/. [Accessed: 24-Nov-2025]
[9] Google, "What is Retrieval-Augmented Generation (RAG)?" [Online]. Available: https://cloud.google.com/use-cases/retrieval-augmented-generation?hl=en. [Accessed: 24-Nov-2025]
[10] OpenAI, "OpenAI Platform API Reference Introduction." [Online]. Available: https://platform.openai.com/docs/api-reference/introduction. [Accessed: 22-Nov-2025]
[11] Roman Markewich, "Adding Native MCP to LlamaIndex Docs," 31-Oct-2025. [Online]. Available: https://www.llamaindex.ai/blog/adding-native-mcp-to-llamaindex-docs. [Accessed: 22-Nov-2025]
[12] Anthropic, "What is the Model Context Protocol (MCP)?," 25-Nov-2024. [Online]. Available: https://modelcontextprotocol.io/docs/getting-started/intro. [Accessed: 22-Nov-2025]
[13] UC Irvine, "144 Time Series Datasets (October 2025) from UC Irvine Machine Learning Repository." [Online]. Available: https://archive.ics.uci.edu/datasets/?skip=0&take=10&sort=desc&orderBy=NumHits&search=&Types=Time-Series. [Accessed: 22-Nov-2025]
[14] MLCommons, University of Virginia, "Time Series Resources Summary Table," 01-Aug-2025. [Online]. Available: https://docs.google.com/document/d/1jNViEKDX_c3Em3MuqLNfdfuqI1_5h97sWuXRZCQ8uT4/edit?tab=t.0. [Accessed: 14-Sep-2025]
[15] Datasets Working Group, "MLCommons Croissant Dataset Schema." [Online]. Available: https://github.com/mlcommons/croissant. [Accessed: 01-Sep-2023]
[16] Jordan Walke, Facebook, "React: The library for web and native user interface," 2011. [Online]. Available: https://react.dev/. [Accessed: 25-Nov-2025]
[17] Kavli Institute for Theoretical Physics, "KITP GenAI25 Meeting: Catalog and Review of Presentations," 04-Jan-2026. [Online]. Available: https://docs.google.com/document/d/1yc1R08Rk4M7HCaIuBEVD5PbWo-26p6Z7tYR9ooeVDLw/edit?usp=sharing. [Accessed: 04-Jan-2026]
[18] A. Radford, J. W. Kim, T. Xu, G. Brockman, C. McLeavey, and I. Sutskever, "Robust speech recognition via large-scale weak supervision," *arXiv [eess.AS]*, pp. 28492–28518, 06-Dec-2022 [Online]. Available: http://arxiv.org/abs/2212.04356
[19] Gemini Team, P. Georgiev, V. I. Lei, R. Burnell, L. Bai, A. Gulati, G. Tanzer, D. Vincent, Z. Pan, S. Wang, S. Mariooryad, Y. Ding, X. Geng, F. Alcober, R. Frostig, *et al.*, "Gemini 1.5: Unlocking multimodal understanding across millions of tokens of context," *arXiv [cs.CL]*, 08-Mar-2024 [Online]. Available: http://arxiv.org/abs/2403.05530
[20] OpenAI, A. Hurst, A. Lerer, A. P. Goucher, A. Perelman, A. Ramesh, A. Clark, A. J. Ostrow, A. Welihinda, A. Hayes, A. Radford, A. Mądry, A. Baker-Whitcomb, A. Beutel, A. Borzunov, *et al.*, "GPT-4o System Card," *arXiv [cs.CL]*, 25-Oct-2024 [Online]. Available: http://arxiv.org/abs/2410.21276
[21] P. Lewis, E. Perez, A. Piktus, F. Petroni, V. Karpukhin, N. Goyal, H. Küttler, M. Lewis, W.-T. Yih, T. Rocktäschel, S. Riedel, and D. Kiela, "Retrieval-augmented generation for knowledge-intensive NLP tasks," *arXiv [cs.CL]*, pp. 9459–9474, 22-May-2020 [Online]. Available: http://arxiv.org/abs/2005.11401
[22] T. Richards, "Auto-GPT: An Autonomous GPT-4 Experiment." [Online]. Available: https://github.com/DataBassGit/Auto-GPT. [Accessed: 21-Mar-2026]
[23] Scott Wu, cognition AI, "Introducing Devin, the first AI software engineer," 12-Mar-2024. [Online]. Available: https://cognition.ai/blog/introducing-devin. [Accessed: 21-Mar-2026]
[24] LangChain, Inc., "LangGraph: Multi-Agent Workflows." [Online]. Available: https://docs.langchain.com/oss/python/langgraph/overview. [Accessed: 21-Mar-2026]
[25] L. Wang, C. Ma, X. Feng, Z. Zhang, H. Yang, J. Zhang, Z. Chen, J. Tang, X. Chen, Y. Lin, W. X. Zhao, Z. Wei, and J. Wen, "A survey on large language model based autonomous agents," *Front. Comput. Sci.*, vol. 18, no. 6, Dec. 2024 [Online]. Available: http://dx.doi.org/10.1007/s11704-024-40231-1
[26] Z. Ji, N. Lee, R. Frieske, T. Yu, D. Su, Y. Xu, E. Ishii, Y. Bang, A. Madotto, and P. Fung, "Survey of hallucination in natural Language Generation," *ACM Comput. Surv.*, Nov. 2022 [Online]. Available:

http://dx.doi.org/10.1145/3571730
[27] Q. Wu, G. Bansal, J. Zhang, Y. Wu, B. Li, E. Zhu, L. Jiang, X. Zhang, S. Zhang, J. Liu, A. H. Awadallah, R. W. White, D. Burger, and C. Wang, “AutoGen: Enabling Next-Gen LLM Applications via Multi-Agent Conversation,” *arXiv [cs.AI]*, 16-Aug-2023 [Online]. Available: http://arxiv.org/abs/2308.08155
[28] Y. Shao, Y. Jiang, T. Kanell, P. Xu, O. Khattab, and M. Lam, “Assisting in writing Wikipedia-like articles from scratch with large language models,” in *Proceedings of the 2024 Conference of the North American Chapter of the Association for Computational Linguistics: Human Language Technologies (Volume 1: Long Papers)*, Mexico City, Mexico, 2024, pp. 6252–6278 [Online]. Available: http://dx.doi.org/10.18653/v1/2024.naacl-long.347
[29] Elicit, “AI for Scientific Research: Elicit helps researchers be 10x more evidence-based.” [Online]. Available: https://elicit.com/. [Accessed: 21-Mar-2026]
[30] Typeset, “SciSpace: Copilot for Research.” [Online]. Available: https://scispace.com/. [Accessed: 21-Mar-2026]
[31] M. Perini, M. P. Pavesi, and L. Attanasio, ScrapeGraphAI, “ScrapeGraphAI: A Web Scraping Python Library based on LLMs.” [Online]. Available: https://github.com/ScrapeGraphAI/Scrapegraph-ai. [Accessed: 21-Mar-2026]
[32] X. Deng, Y. Gu, B. Zheng, S. Chen, S. Stevens, B. Wang, H. Sun, and Y. Su, “Mind2Web: Towards a generalist agent for the web,” *arXiv [cs.CL]*, pp. 28091–28114, 09-Jun-2023 [Online]. Available: http://arxiv.org/abs/2306.06070
[33] M. Bishop and M. Dilger, “Checking for race conditions in file accesses,” *Computing systems*, vol. 2, no. 2, pp. 131–152, 1996 [Online]. Available: https://www.usenix.org/legacy/publications/compsystems/1996/spr_bishop.pdf
[34] Z. Jiang, F. F. Xu, J. Araki, and G. Neubig, “How can we know what language models know?,” *Trans. Assoc. Comput. Linguist.*, vol. 8, pp. 423–438, Dec. 2020 [Online]. Available: http://dx.doi.org/10.1162/tacl_a_00324/96460
[35] R. Fielding, Y. Lafon, and J. Reschke, “Hypertext transfer protocol (HTTP/1.1): range requests,” 2014 [Online]. Available: https://www.rfc-editor.org/rfc/rfc7233.html
[36] P. Christen, “The data matching process,” in *Data Matching*, Berlin, Heidelberg: Springer Berlin Heidelberg, 2012, pp. 23–35 [Online]. Available: http://dx.doi.org/10.1007/978-3-642-31164-2_2
[37] N. Shinn, F. Cassano, E. Berman, A. Gopinath, K. Narasimhan, and S. Yao, “Reflexion: Language agents with verbal reinforcement learning,” *arXiv [cs.AI]*, pp. 8634–8652, 20-Mar-2023 [Online]. Available: http://arxiv.org/abs/2303.11366
[38] H. Wang, T. Fu, Y. Du, W. Gao, K. Huang, Z. Liu, P. Chandak, S. Liu, P. Van Katwyk, A. Deac, A. Anandkumar, K. Bergen, C. P. Gomes, S. Ho, P. Kohli, *et al.*, “Scientific discovery in the age of artificial intelligence,” *Nature*, vol. 620, no. 7972, pp. 47–60, Aug. 2023 [Online]. Available: http://dx.doi.org/10.1038/s41586-023-06221-2
[39] T. Wolf, L. Debut, V. Sanh, J. Chaumond, C. Delangue, A. Moi, P. Cistac, T. Rault, R. Louf, M. Funtowicz, J. Davison, S. Shleifer, P. von Platen, C. Ma, Y. Jernite, *et al.*, “Transformers: State-of-the-art natural language processing,” in *Proceedings of the 2020 Conference on Empirical Methods in Natural Language Processing: System Demonstrations*, Online, 2020, pp. 38–45 [Online]. Available: http://dx.doi.org/10.18653/v1/2020.emnlp-demos.6
[40] I. D. Baxter, A. Yahin, L. Moura, M. Sant’Anna, and L. Bier, “Clone detection using abstract syntax trees,” in *Proceedings. International Conference on Software Maintenance (Cat. No. 98CB36272)*, Bethesda, MD, USA, 2002, pp. 368–377 [Online]. Available: http://dx.doi.org/10.1109/ICSM.1998.738528
[41] C. Lattner and V. Adve, “LLVM: A compilation framework for lifelong program analysis & transformation,” in *International Symposium on Code Generation and Optimization, 2004. CGO 2004*, San Jose, CA, USA, 2004, pp. 75–86 [Online]. Available: http://dx.doi.org/10.1109/CGO.2004.1281665
[42] B. Tran, M. Maiterth, W. Shin, M. D. Sinclair, and S. Venkataraman, “Wattchmen: Watching the wattchers -- high fidelity, flexible GPU energy modeling,” *arXiv [cs.AR]*, 27-Mar-2026 [Online]. Available: http://arxiv.org/abs/2603.26435
[43] B. Hawks, G. von Laszewski, M. D. Sinclair, M. Colombo, S. Venkataraman, R. Jain, Y. Jiang, N.

Tran, and G. Fox, “An MLCommons Scientific Benchmarks Ontology,” *arXiv [cs.LG]*, 06-Nov-2025 [Online]. Available: http://dx.doi.org/10.48550/arXiv.2511.05614
[44] F. Fahim, B. Hawks, C. Herwig, J. Hirschauer, S. Jindariani, N. Tran, L. P. Carloni, G. Di Guglielmo, P. Harris, J. Krupa, D. Rankin, M. B. Valentin, J. Hester, Y. Luo, J. Mamish, *et al.*, “Hls4ml: An open-source codesign workflow to empower scientific low-power machine learning devices,” *arXiv [cs.LG]*, 09-Mar-2021 [Online]. Available: http://arxiv.org/abs/2103.05579
[45] “Python Colab code: Ultimate Science Summarizer43-Spriggs,” *{Geoffrey Fox}*, 03-Jan-2026. [Online]. Available: https://colab.research.google.com/drive/1BgsPYcLDYbjwxyvrqb3YXJHxl3bOImML?usp=sharing. [Accessed: 04-Jan-2026]
[46] J. Z. Tam, P. Grosset, D. Banesh, N. Ramachandra, T. L. Turton, and J. Ahrens, “InferA: A Smart Assistant for Cosmological Ensemble Data,” in *Proceedings of the SC ’25 Workshops of the International Conference for High Performance Computing, Networking, Storage and Analysis*, St Louis MO USA, 2025, pp. 20–28 [Online]. Available: http://dx.doi.org/10.1145/3731599.3767342
[47] Tam, Justin Z., Pascal Grosset, Divya Banesh, Nesar Ramachandra, Terece L. Turton, and James Ahrens, “GitHub for InferA: A Smart Assistant for Cosmological Ensemble Data,” 2025. [Online]. Available: https://github.com/lanl/InferA. [Accessed: 29-May-2026]
[48] M. Grosskopf, N. Debardeleben, R. Bent, R. Somasundaram, I. Michaud, A. Lui, A. Wadell, W. D. Graham, G. A. Wimmer, S. Shivakumar, J. V. Gallart, H. Nagarajan, and E. Lawrence, “URSA: The universal research and scientific agent,” *arXiv [cs.AI]*, 27-Jun-2025 [Online]. Available: http://arxiv.org/abs/2506.22653
[49] T. J. Hobbs, S. Das Bakshi, P. Barry, C. Bissolotti, I. Cloet, S. Corrodi, Z. Djurcic, S. Habib, K. Heitmann, W. Hopkins, S. Joosten, B. Kriesten, N. Ramachandra, S. Roy, A. Wells, *et al.*, “ArgoLOOM: agentic AI for fundamental physics from quarks to cosmos,” *arXiv [hep-ph]*, 2026 [Online]. Available: http://arxiv.org/abs/2510.02426
[50] Z. Ziheng, H. Tang, J. Zhang, H. Lin, B. Yang, Q. Long, F. Sun, Y. Sun, Y. Liang, Y. N. Wu, D. Terzopoulos, and X. Gao, “Can current agents close the discovery-to-application gap? A case study in Minecraft,” *arXiv [cs.AI]*, 27-Apr-2026 [Online]. Available: http://arxiv.org/abs/2604.24697
[51] X. Zhuang, C. Zhou, K. Feng, Z. Zhu, Y. Gao, Y. Zhong, Y. Zhang, J. Huang, K. Ding, L. Bai, H. Wang, Q. Zhang, and H. Chen, “Embodied science: Closing the discovery loop with agentic embodied AI,” *arXiv [cs.AI]*, 20-Mar-2026 [Online]. Available: http://arxiv.org/abs/2603.19782
[52] M. Roy and S. Roy, “Neuro-Symbolic Hypothesis Engine: A Unified Architecture for Autonomous Scientific Hypothesis Generation,” 2025 [Online]. Available: https://openreview.net/pdf?id=bXgeIK17S7
[53] S. Xia, Y. Sun, and P. Liu, “SR-scientist: Scientific equation discovery with agentic AI,” *arXiv [cs.AI]*, 13-Oct-2025 [Online]. Available: http://arxiv.org/abs/2510.11661
[54] E. A. Moreno, S. Bright-Thonney, A. Novak, D. Garcia, and P. Harris, “AI agents can already autonomously perform experimental high energy physics,” *arXiv [hep-ex]*, 20-Mar-2026 [Online]. Available: http://arxiv.org/abs/2603.20179
[55] F. Cappello, S. Madireddy, R. Underwood, N. Getty, N. L.-P. Chia, N. Ramachandra, J. Nguyen, M. Keceli, T. Mallick, Z. Li, M. Ngom, C. Zhang, A. Yanguas-Gil, E. Antoniuk, B. Kailkhura, *et al.*, “EAIRA: Establishing a methodology for Evaluating AI models as scientific Research Assistants,” *arXiv [cs.AI]*, 27-Feb-2025 [Online]. Available: http://arxiv.org/abs/2502.20309
[56] Michael J. Sullivan, “Knowledge Graph Modeling: Time series micro-pattern using GIST,” 21-Jul-2022. [Online]. Available: https://www.ateam-oracle.com/post/knowledge-graph-modeling-time-series-micro-pattern-using-gist. [Accessed: 24-Nov-2025]
[57] Mike Dillinger, “Continuous Knowledge Graphs for Neurosymbolic AI,” 09-Apr-2024. [Online]. Available: https://medium.com/@mike.dillinger/continuous-knowledge-graphs-for-neurosymbolic-ai-9000b4aa0eb7. [Accessed: 24-Nov-2025]
[58] Noah Mayerhofer, Neo4j, “How to Convert Unstructured Text to Knowledge Graphs Using LLMs,” 30-Jul-2025. [Online]. Available: https://neo4j.com/blog/developer/unstructured-text-to-knowledge-graph/. [Accessed: 24-Nov-2025]

[59] Rahul Kumar, “Beyond LLMs: Building a Graph-RAG Agentic Architecture for 70% Faster ECM Automation,” 04-Nov-2025. [Online]. Available: https://medium.com/@hellorahulk/beyond-llms-building-a-graph-rag-agentic-architecture-for-70-faster-ecm-automation-299b05d026fb. [Accessed: 24-Nov-2025]
[60] Subrata Samanta, “Hybrid Graph RAG: Harnessing Graph and Vector Databases for Advanced 10-K Insights,” 01-May-2025. [Online]. Available: https://medium.com/@subrata-samanta/hybrid-graph-rag-harnessing-graph-and-vector-for-financial-analysis-72c3a9f1a09d. [Accessed: 24-Nov-2025]
[61] sara Tilly, “HybridRAG and Why Combine Vector Embeddings with Knowledge Graphs for RAG?,” 03-Sep-2025. [Online]. Available: https://memgraph.com/blog/why-hybridrag. [Accessed: 24-Nov-2025]
[62] Denise Gosnell, Vivien de Saint Pern, “Improving Retrieval Augmented Generation accuracy with GraphRAG,” 23-Dec-2024. [Online]. Available: https://aws.amazon.com/blogs/machine-learning/improving-retrieval-augmented-generation-accuracy-with-graphrag/. [Accessed: 24-Nov-2025]
[63] B. Peng, Y. Zhu, Y. Liu, X. Bo, H. Shi, C. Hong, Y. Zhang, and S. Tang, “Graph retrieval-Augmented Generation: A Survey,” *ACM Trans. Inf. Syst.*, no. 3777378, Nov. 2025 [Online]. Available: http://dx.doi.org/10.1145/3777378
[64] IBM, “What is GraphRAG?” [Online]. Available: https://www.ibm.com/think/topics/graphrag. [Accessed: 24-Nov-2025]
[65] D. Edge, H. Trinh, N. Cheng, J. Bradley, A. Chao, A. Mody, S. Truitt, D. Metropolitansky, R. O. Ness, and J. Larson, “From local to global: A graph RAG approach to query-focused summarization,” *arXiv [cs.CL]*, 24-Apr-2024 [Online]. Available: http://arxiv.org/abs/2404.16130
[66] Bivek Ranuji, “Vector RAG & Graph RAG: A Quick Read on Where to Use What,” 10-Nov-2025. [Online]. Available: https://bivekrenuji.medium.com/vector-rag-graph-rag-a-quick-read-on-where-to-use-what-9d14a0020731. [Accessed: 24-Nov-2025]
[67] Microsoft, “Project GraphRAG: LLM-Derived Knowledge Graphs,” 2024. [Online]. Available: https://www.microsoft.com/en-us/research/project/graphrag/. [Accessed: 24-Nov-2025]
[68] W. Wei, T. Yang, H. Chen, R. A. Rossi, Y. Zhao, F. Dernoncourt, and H. Eldardiry, “Efficient model selection for time series forecasting via LLMs,” *arXiv [cs.LG]*, 02-Apr-2025 [Online]. Available: http://arxiv.org/abs/2504.02119
[69] N. Xu, C. Kosma, and M. Vazirgiannis, “TimeGNN: Temporal dynamic graph learning for time series forecasting,” *arXiv [cs.LG]*, 27-Jul-2023 [Online]. Available: http://dx.doi.org/10.1007/978-3-031-53468-3_8
[70] W. Zhang, Y. Bei, L. Yang, H. P. Zou, P. Zhou, A. Liu, Y. Li, H. Chen, J. Wang, Y. Wang, F. Huang, S. Zhou, J. Bu, A. Lin, J. Caverlee, *et al.*, “Cold-start recommendation towards the era of large language models (LLMs): A comprehensive survey and roadmap,” *arXiv [cs.IR]*, 03-Jan-2025 [Online]. Available: http://arxiv.org/abs/2501.01945
[71] “OpenML: A worldwide machine learning lab.” [Online]. Available: https://www.openml.org/. [Accessed: 10-Sep-2023]
[72] J. Vanschoren, J. N. van Rijn, B. Bischl, and L. Torgo, “OpenML: networked science in machine learning,” *SIGKDD Explor.*, vol. 15, no. 2, pp. 49–60, Jun. 2014 [Online]. Available: http://dx.doi.org/10.1145/2641190.2641198
[73] A. Paleyes, R.-G. Urma, and N. D. Lawrence, “Challenges in deploying machine learning: A survey of case studies,” *ACM Comput. Surv.*, vol. 55, no. 6, pp. 1–29, Jul. 2023 [Online]. Available: http://dx.doi.org/10.1145/3533378
[74] F. Hutter, L. Kotthoff, and J. Vanschoren, *Automated machine learning: Methods, systems, challenges*. Springer Nature, 2019 [Online]. Available: http://dx.doi.org/10.1007/978-3-030-05318-5
[75] A. Alsharef, K. Aggarwal, Sonia, M. Kumar, and A. Mishra, “Review of ML and AutoML solutions to forecast time-series data,” *Arch. Comput. Methods Eng.*, vol. 29, no. 7, pp. 5297–5311, Jun. 2022 [Online]. Available: http://dx.doi.org/10.1007/s11831-022-09765-0
[76] B. Bohnet, V. Q. Tran, P. Verga, R. Aharoni, D. Andor, L. B. Soares, M. Ciaramita, J. Eisenstein, K. Ganchev, J. Herzig, K. Hui, T. Kwiatkowski, J. Ma, J. Ni, L. S. Saralegui, *et al.*, “Attributed question

answering: Evaluation and modeling for attributed large language models," *arXiv [cs.CL]*, 15-Dec-2022 [Online]. Available: http://arxiv.org/abs/2212.08037
[77] S. Kadavath, T. Conerly, A. Askell, T. Henighan, D. Drain, E. Perez, N. Schiefer, Z. Hatfield-Dodds, N. DasSarma, E. Tran-Johnson, S. Johnston, S. El-Showk, A. Jones, N. Elhage, T. Hume, *et al.*, "Language models (mostly) know what they know," *arXiv [cs.CL]*, 11-Jul-2022 [Online]. Available: http://arxiv.org/abs/2207.05221
[78] "MLCube: MLCommons Interoperable Container interface for Machine Learning." [Online]. Available: https://mlcommons.org/en/mlcube/. [Accessed: 10-May-2022]
[79] X. Wang, K. Smith-Miles, and R. Hyndman, "Rule induction for forecasting method selection: Meta-learning the characteristics of univariate time series," *Neurocomputing*, vol. 72, no. 10–12, pp. 2581–2594, Jun. 2009 [Online]. Available: http://dx.doi.org/10.1016/j.neucom.2008.10.017
[80] T. S. Talagala, R. J. Hyndman, and G. Athanasopoulos, "Meta‐learning how to forecast time series," *J. Forecast.*, vol. 42, no. 6, pp. 1476–1501, Sep. 2023 [Online]. Available: http://dx.doi.org/10.1002/for.2963
[81] Judy Fox, "Prototype TimeSeries Commons Website." [Online]. Available: https://uva-mlsys.github.io/Time-Series-Commons. [Accessed: 30-Nov-2025]
[82] John Hopkins, MGHPCC, NCSA, PSC, renci, SDSC, "Open Storage Network." [Online]. Available: https://www.openstoragenetwork.org/. [Accessed: 26-Nov-2025]
[83] Luke Emberson, Epoch AI, "Open-weight models lag state-of-the-art by around 3 months on average," 30-Oct-2025. [Online]. Available: https://epoch.ai/data-insights/open-weights-vs-closed-weights-models. [Accessed: 29-Nov-2025]
[84] Venkat Somala, Luke Emberson, Epoch AI, "Frontier AI performance becomes accessible on consumer hardware within a year," 15-Aug-2025. [Online]. Available: https://epoch.ai/data-insights/consumer-gpu-model-gap. [Accessed: 29-Nov-2025]
[85] Mallory Mejias, Sidecar.ai, "Why AI Is About to Become (Almost) Free - And What That Means for Your Association," 15-Oct-2025. [Online]. Available: https://sidecar.ai/blog/why-ai-is-about-to-become-almost-free-and-what-that-means-for-your-association. [Accessed: 29-Nov-2025]
[86] Stanford University Human Centered Artificial Intelligence, "AI Reshapes Global Power: Insights from Stanford HAI's Congressional Boot Camp," 02-Sep-2025. [Online]. Available: https://hai.stanford.edu/news/ai-reshapes-global-power-insights-from-stanford-hais-congressional-boot-camp. [Accessed: 29-Nov-2025]
[87] Markus Heiser, Adam Tauber, "SearXNG metasearch engine GitHub," 2021. [Online]. Available: https://github.com/searxng/searxng. [Accessed: 18-May-2026]
[88] Google DeepMind, "Gemma Open LLM model," 2026. [Online]. Available: https://deepmind.google/models/gemma/. [Accessed: 18-May-2026]
[89] K. Greshake, S. Abdelnabi, S. Mishra, C. Endres, T. Holz, and M. Fritz, "Not what you've signed up for: Compromising real-world LLM-integrated applications with indirect prompt injection," in *Proceedings of the 16th ACM Workshop on Artificial Intelligence and Security*, Copenhagen Denmark, 2023, pp. 79–90 [Online]. Available: https://dl.acm.org/doi/abs/10.1145/3605764.3623985
[90] K. Greshake, S. Abdelnabi, S. Mishra, C. Endres, T. Holz, and M. Fritz, "More than you've asked for: A comprehensive analysis of novel prompt injection threats to application-integrated large language models," *arXiv preprint arXiv:2302. 12173*, vol. 27, 2023 [Online]. Available: https://scholar.google.com/citations?user=JGbEWeAAAAAJ&hl=en&oi=sra
[91] F. Perez and I. Ribeiro, "Ignore previous prompt: Attack techniques for language models," *arXiv [cs.CL]*, 17-Nov-2022 [Online]. Available: http://arxiv.org/abs/2211.09527
[92] The OWASP Foundation, "OWASP Top 10 for Large Language Model Applications." [Online]. Available: https://owasp.org/www-project-top-10-for-large-language-model-applications/. [Accessed: 29-Nov-2025]
[93] H. Inan, K. Upasani, J. Chi, R. Rungta, K. Iyer, Y. Mao, M. Tontchev, Q. Hu, B. Fuller, D. Testuggine, and M. Khabsa, "Llama Guard: LLM-based input-output safeguard for Human-AI conversations," *arXiv [cs.CL]*, 07-Dec-2023 [Online]. Available: http://arxiv.org/abs/2312.06674
[94] NVIDIA, "NeMo Guardrails: A toolkit for controllable and safe LLM-powered applications." [Online]. Available: https://github.com/NVIDIA-NeMo/Guardrails. [Accessed: 29-Nov-2025]

[95] Guardrails AI, "Validators: Guardrails components that are used to validate an aspect of an LLM workflow." [Online]. Available: https://guardrailsai.com/hub. [Accessed: 29-Nov-2025]
[96] Pydantic, "Pydantic Validation: the most widely used data validation library for Python." [Online]. Available: https://docs.pydantic.dev/latest/. [Accessed: 29-Nov-2025]
[97] Google, "*gVisor*: open-source Linux-compatible sandbox. The Container Security Platform." [Online]. Available: https://gvisor.dev/. [Accessed: 29-Nov-2025]
[98] OpenInfra Foundation, "Kata Containers: The speed of containers, the security of VMs." [Online]. Available: https://katacontainers.io/. [Accessed: 29-Nov-2025]
[99] Y. Liu, G. Qin, Z. Shi, Z. Chen, C. Yang, X. Huang, J. Wang, and M. Long, "Sundial: A family of highly capable time series foundation models," *arXiv [cs.LG]*, 02-Feb-2025 [Online]. Available: http://arxiv.org/abs/2502.00816
[100] G. von Laszewski, P. Luszczek, W. Brewer, J. Thiyagalingam, J. Papay, G. C. Fox, A. Foundjem, G. Barrett, M. Emani, S. V. Moore, V. J. Reddi, G. Farrell, M. D. Sinclair, C. Kirkpatrick, S. Venkataraman, *et al.*, "AI Benchmarks Carpentry and Democratization." Biocomplexity Institute, University of Virginia, Charlottesville, VA, USA, 2025 [Online]. Available: https://www.overleaf.com/read/xrysvzdnyjgt#a2ff11
[101] Geoffrey Fox, "Python Colab Code for Deep Collector March 2026," 21-Mar-2026. [Online]. Available: https://colab.research.google.com/drive/1WA0Zs4nDvPjFJGArmjThhzzBL_ToB1xX?usp=sharing. [Accessed: 22-Mar-2026]

# Appendix A: Engineering Details of DeepCollector

## A.1 Introduction

While the main body of this paper focuses on the theoretical methodology and resulting dataset catalog produced by DeepCollector, deploying an autonomous, multi-agent Large Language Model (LLM) pipeline in practice requires overcoming significant systems engineering challenges. Moving from a conceptual proof-of-concept to a robust, production-ready extraction engine necessitates complex state management, dynamic rate-limit handling, hallucination mitigation, and strict schema enforcement. This appendix details the technical architecture, operational modes, and empirical resource utilization of the DeepCollector software framework, providing a transparent view of the computational costs and engineering infrastructure required to automate the curation of a large-scale knowledge graph. The Colab code can be accessed [101].

## A.2 System Overview and Execution Modes

DeepCollector is an autonomous, agentic pipeline designed to compile, standardize, and maintain a comprehensive knowledge graph of time-series datasets. It utilizes a hybrid deterministic/probabilistic architecture to ingest data from academic papers, GitHub repositories, and structured APIs into a centralized Google Sheets database.

The pipeline is governed by a global configuration parameter, MODE. Users execute batches by defining a list of target projects and assigning the desired operational mode:

- **PLAN:** A diagnostic mode that scans the current state of the Knowledge Base against the Master Project Sheet, identifying un-run projects, projects requiring repair due to missing mandatory fields, and overlapping runs requiring a merge.
- **AGENT:** The primary ingestion mode. Utilizing Deep Research and Cellular RAG, it discovers novel datasets from unstructured source material, grounds the data locally, and extracts required schema fields.
- **HARVEST:** The deterministic bypass used for structured repositories (e.g., UCI). It ignores LLM heuristics in favor of strict HTML/API scraping to guarantee exact schema conformity and 100% confidence scores.
- **REPAIR:** A targeted surgical mode that isolates specific datasets missing mandatory schema fields (like physical download URLs) and deploys targeted web searches to fill the gaps without overwriting existing high-confidence data.
- **MERGE:** The deduplication engine. Invokes the LLM Arbitrator to evaluate fuzzy matches across the database, consolidating duplicate dataset entries and pooling their project citations into a single Golden Record.
- **REVIEW:** The Quality Auditor. It scans the database against strict schema definitions, permanently quarantining any datasets that exhibit low confidence scores (< 0.5) or missing required source links.

## A.3 Software Architecture and Component Modularity

Based on a highly modular object-oriented architecture, the DeepCollector application represents a substantive software engineering effort. To manage the robust error-handling, asynchronous API polling, and complex prompt engineering required for autonomous operation, the codebase is partitioned into distinct functional domains.

**Categorization & Functional Breakdown:**

- **Agent Core & State Engine**

- *Modules:* `core/agent.py`, `core/rag_engine.py`, `core/state.py`
- *Purpose:* The brain of the system. Manages the phase-based execution loop (Bootstrapping, Discovery, Grounding, Extraction), the Tri-State telemetry (measuring net-new vs. confirmed data), and the "Cellular RAG" logic that dynamically generates sub-queries to isolate inference to specific missing database fields.

- **Knowledge Base Management & Deduplication**
  - *Modules:* `kb/manager.py`, `kb/merger.py`, `kb/quality.py`, `kb/maintenance.py`
  - *Purpose:* Handles complex concurrent locking, reading, and JSON-compliant batch-writing to the Google Sheets API. It houses the LLM-powered fuzzy deduplication engine and the strict Quality Auditor schema-enforcement logic.
- **Configuration, Context & Utilities**
  - *Modules:* `config/settings.py`, `utils/project_loader.py`, etc.
  - *Purpose:* Stores the strict schema definitions, manages dual-authentication (Google Drive & Sheets), provides detailed time-profiling and Tri-State analytics for academic benchmarking, and dynamically loads pipeline configurations from the Master Google Sheet.
- **Tooling & Deep Research**
  - *Modules:* `tools/research.py`, `tools/ddi.py`
  - *Purpose:* Bridges the LLM to the external world. Implements the Dynamic Data Investigator (DDI) for surgical data-file inspection, connects to the LlamaIndex framework for handling unstructured context, and manages the asynchronous, quota-aware polling loop for the Deep Research agent.
- **Deterministic Harvesting**
  - *Modules:* `harvesting/uci_harvester.py`, `harvesting/base_harvester.py`
  - *Purpose:* Specialized, hard-coded web scrapers that bypass probabilistic LLM guessing for structured targets like the UCI Machine Learning Repository, guaranteeing exact schema conformity and high-confidence links.

## A.4 Telemetry and Data Confidence

To ensure data integrity, all fields populated by DeepCollector are accompanied by a Confidence Score (0.0 to 1.0) and Telemetry Context or provenance explaining how the agent arrived at the value. This transparency allows researchers to track provenance across automated extractions. Users are advised to prioritize manual verification on any fields returning scores below the 0.80 threshold.